\documentclass{article}
\usepackage{spconf,amsmath,graphicx}
\usepackage{subfig}
\usepackage{epstopdf}
\usepackage{flushend}

\usepackage{color}
\usepackage[hyphens]{url}

\title{FREEHAND SKETCH RECOGNITION USING DEEP FEATURES}

\name{Ravi Kiran Sarvadevabhatla,R. Venkatesh Babu}
\address{Video Analytics Lab, SERC,\\Indian Institute of Science, Bangalore, India.\\ ravikiran@ssl.serc.iisc.in,venky@serc.iisc.in}

\begin{document} 
\ninept 
\maketitle

\begin{abstract}
Freehand sketches often contain sparse visual detail. In spite of the sparsity, they are easily and consistently recognized by humans across cultures, languages and age groups. Therefore, analyzing such sparse sketches can aid our understanding of the neuro-cognitive processes involved in visual representation and recognition. In the recent past, Convolutional Neural Networks (CNNs) have emerged as a powerful framework for feature representation and recognition for a variety of image domains. However, the domain of sketch images has not been explored. This paper introduces a freehand sketch recognition framework based on ``deep" features extracted from CNNs. We use two popular CNNs for our experiments -- Imagenet CNN and a modified version of LeNet CNN. We evaluate our recognition framework on a publicly available benchmark database containing thousands of freehand sketches depicting everyday objects. Our results are an improvement over the existing state-of-the-art accuracies by $3\% - 11\%$. The effectiveness and relative compactness of our deep features also make them an ideal candidate for related problems such as sketch-based image retrieval. In addition, we provide a preliminary glimpse of how such features can help identify relative importance of crucial attributes (e.g. object-parts) in the sketched objects.
\end{abstract}


\section{Introduction}
\begin{figure}[ht]
\centering
\includegraphics[width=.25\linewidth]{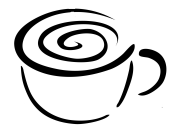}
\caption{In spite of minimal detail, we can recognize the line sketch easily and correctly as belonging to the category \texttt{cup}.}
\label{fig:cup}
\end{figure}

A fascinating spectrum  -- from realistic depictions to sparsely drawn sketches -- exists among categories selected as subjects for hand-drawn art. In particular, consider the category of common everyday objects. An instance of such a sketch can be seen in Figure \ref{fig:cup}. Though containing minimal detail, the object category to which it belongs is easily determined. This suggests an inherent sparseness in the human neuro-visual representation of the object. Therefore, studying such sparse sketches can aid our understanding of the cognitive processes involved and spur the design of efficient visual classifiers. Another important reason for studying such sketches is the fact that they form a universal language -- the underlying subject of interest can usually be identified correctly, overcoming barriers of culture, language, time period and age group. Consequently, studying sketches can help identify factors that contribute to such universality among human kind. 

Freehand sketches are typically formed as a composition of primitive hand-drawn curves (called strokes) added sequentially over time.
Deciphering freehand sketches can be viewed under the lens of image category recognition, a well studied problem in the computer vision community. In the recent years, deep-learning frameworks based on Convolutional Neural Networks (CNNs)\cite{krizhevsky2012imagenet}\cite{girshick2014rcnn}\cite{DonahueJVHZTD13} have shown impressive performance on challenging image recognition datasets. However, to the best of our knowledge, the domain of freehand sketch images has not been explored in this context.

In this paper, we utilize ``deep" features based on CNNs to recognize hand-drawn sketches across numerous object categories. Our results are an improvement over the existing state-of-the-art accuracies by $3\% - 11\%$. The effectiveness and relative compactness of our features make them an ideal candidate for related problems such as sketch-based image retrieval. In addition, we provide a preliminary glimpse of how such features
can help identify relative importance of crucial attributes (e.g. object-parts) in the sketched objects. More generally, we hope that our work will spur interest in analysis of sketches utilizing cutting-edge tools of the deep learning methodology.

The rest of the paper is organized as follows: We briefly review related literature in Section \ref{sec:relatedwork}. We describe the sketch database and certain issues related to its usage in Section \ref{sec:db}. Section \ref{sec:featex} describes the ``deep" feature extraction process and the recognition mechanism. A comparative evaluation of the recognition framework is presented in \ref{sec:eval}. Section \ref{sec:conclusion} concludes the paper by outlining work in progress and suggesting directions for future work.

 \section{Related Work}
\label{sec:relatedwork}

\begin{figure*}[ht!]
  \includegraphics[height=3.5cm,width=\textwidth]{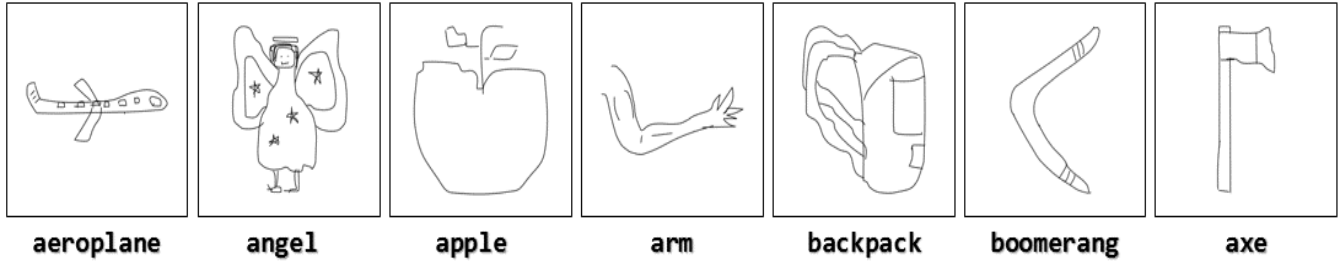}
  \caption{Freehand sketches belonging to various object categories.}
	\label{fig:primal-sketches}
\end{figure*}

A significant body of work has examined freehand line sketches in the context of recognition\cite{KMY06}\cite{QiGLZXS13} and content-based image retrieval problems\cite{Hu:2013:PEG:2479988.2480107}\cite{Hu2010}. To retain focus, we examine literature related to sketch recognition, although some aspects of the retrieval problem, particularly the feature extraction, remain relevant. Methods for freehand sketch recognition exhibit two broad themes -- domain-specific and general. In the former category, recognition systems have been built for sketches related to mathematical expressions\cite{math}, emergency management and military drawings\cite{military} and chemistry diagrams\cite{chem}. These systems tend to make domain-specific assumptions about the structure and syntax of the sketch strokes. In the latter category viz. general, a recent attempt to capture the general nature of freehand sketches was presented by Eitz et al.\cite{eitz} via analysis of a human-drawn sketch database containing $250$ commonly encountered object categories. Eitz et al. also describe a sketch classifier which obtains an average accuracy of $54\%$ across the $250$ categories when $80\%$ of the sketches are used for training. Ros\'{a}lia et al.\cite{rosalia} improve upon this result using a Fisher vector image representation\cite{perronnin2007fisher}, raising the average accuracy to $67\%$. Refer to the works of Eitz et al.\cite{eitz} and Ros\'{a}lia et al.\cite{rosalia} for a more detailed taxonomy of freehand sketch classification techniques, features used and results therein. 

The versatile recognition and representation framework provided by CNNs has resulted in an avalanche of deep learning based works, too numerous and varied to summarize here. For a dynamic, annotated bibliography of work related to CNNs, refer to Amund et al.\cite{amund}. The CNNs used in our experiments -- Imagenet CNN\cite{krizhevsky2012imagenet} and LeNet CNN\cite{lecun1998gradient} -- are in fact, merely two notable examples among the many that have been proposed for problems in computer vision and image processing. To the best of our knowledge, CNNs have not been utilized in the analysis of general freehand sketches. Two works, however, can be considered peripherally related in this context. The first is the work of Fu et al.\cite{luoting} which utilizes CNNs to recognize pre-defined symbols in sketches from different engineering domains. The second work is that of Wan et al.\cite{wen} which attempts to learn an auto-encoder (a deep learning architecture similar to CNNs) to recognize faces from a database of face photos and their sketched versions.
 
\section{The sketch database}
\label{sec:db}

For our experiments, we use the publicly available freehand sketch database of Eitz et al.\cite{eitz}. This database contains a set of $20,000$ hand-drawn sketches evenly distributed across $250$ object categories. These sketches have been obtained by crowdsourcing across the general population. As such, they are a good starting point for analyzing the neuro-cognitive underpinnings of the sketching process by humans. A few examples from the database can be seen in Figure \ref{fig:primal-sketches}. An interesting feature of this database is that the temporal stroke information (the sequential order in which the strokes were drawn) for a sketch has also been provided. 

\begin{figure}[t]
\centering
\includegraphics[width=.65\linewidth]{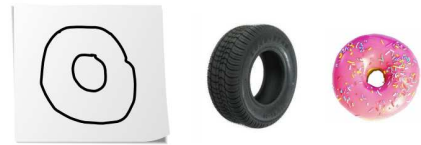}
\caption{An ambiguously drawn sketch (left) with two plausible categorizations - \texttt{tyre} (middle) and \texttt{donut} (right). Figure has been taken from \cite{rosalia}.}
\label{fig:tyredonut}
\end{figure}

While this database is quite comprehensive and general in coverage of object categories, it has its share of shortcomings, a number of which have been analyzed by Ros\'{a}lia et al.\cite{rosalia}. One of the major shortcomings is the presence of ambiguously drawn sketches whose identity is difficult to discern even for fellow human beings (See Figure \ref{fig:tyredonut}. Is the sketch on the left depicted by the two concentric shapes a \texttt{tyre} or a \texttt{donut} ?). To address this situation, Ros\'{a}lia et al. employ a human-evaluation based technique and identify a subset containing $160$ non-ambiguous object categories which can be utilized as a more reliable benchmark database for evaluating sketch recognition systems (Refer to Section $5$ of \cite{rosalia} for details). For our experiments, we utilize the sketches from this curated set of $160$ object categories for analyzing the performance of the sketch recognition system. Furthermore, following \cite{rosalia}, we uniformly consider $56$ sketches from each category.

\section{Feature Extraction and Classification}
\label{sec:featex}

\begin{figure*}[ht!]
\begin{minipage}[b]{0.45\linewidth}
\centering
\includegraphics[width=\textwidth]{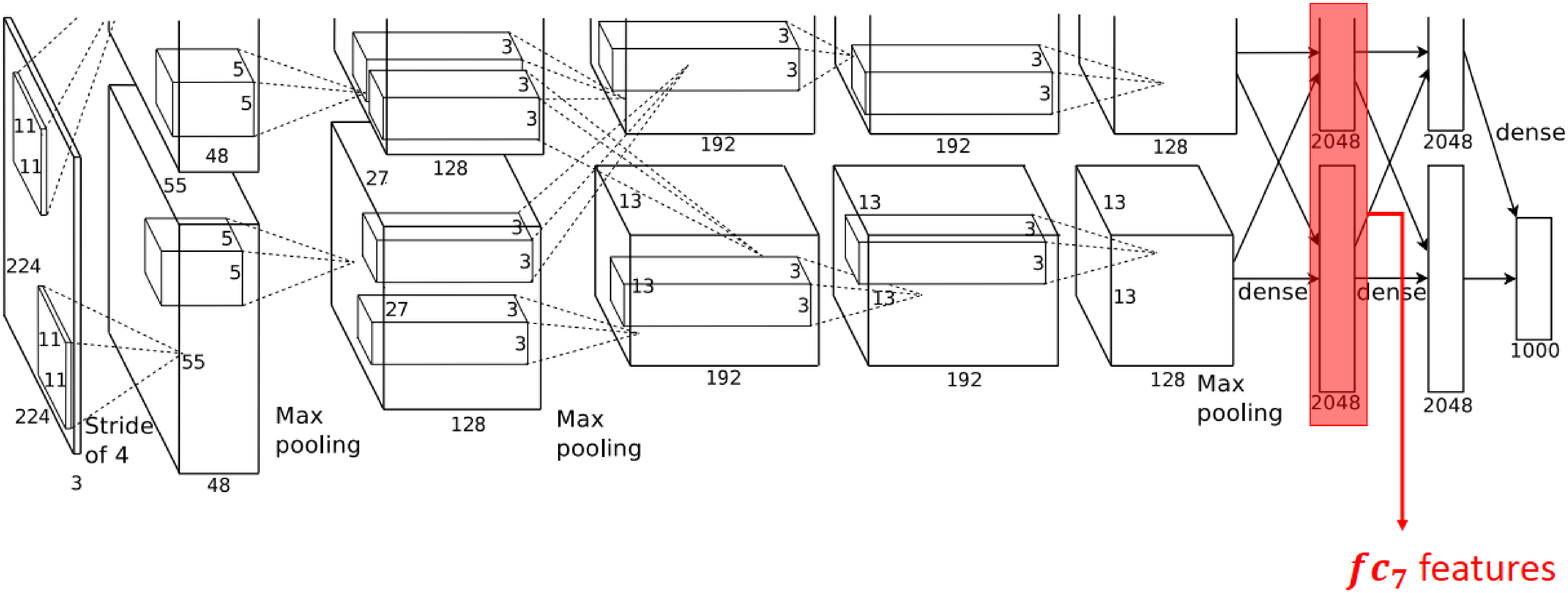}
\caption{Imagenet CNN showing the layer (shaded in red) $fc_7$ from which the $4096$-dimensional sketch features are extracted. A portion of the figure is taken from \cite{krizhevsky2012imagenet}.}
\label{fig:figure1}
\end{minipage}
\hspace{0.5cm}
\begin{minipage}[b]{0.45\linewidth}
\centering
\includegraphics[width=\textwidth]{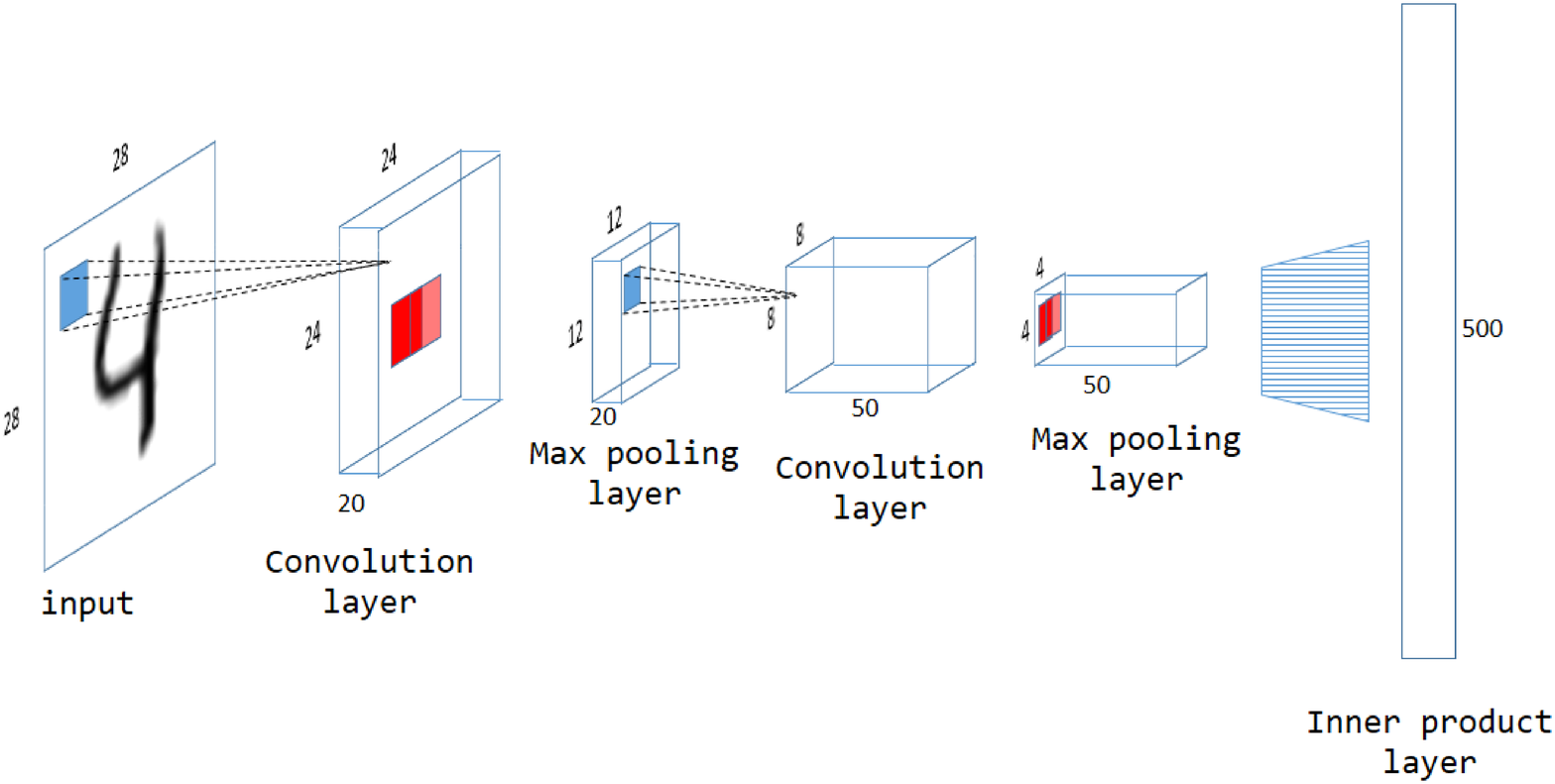}
\caption{Modified version of LeNet CNN used for our experiments (only the first $5$ layers are shown for simplicity). The overlapping boxes shaded in red indicate stride. The hatched region between max pooling layer and inner-product layer $ip_1$ denotes fully connectedness. Note that we get a $500$-dimensional feature vector at the end of the process.}
\label{fig:figure2}
\end{minipage}
\end{figure*}

The sketch features are extracted using pre-trained Convolutional Neural Networks (CNNs). We first provide a brief introduction to CNNs, describe the ones used in our experiments and in the process of this description, provide details of feature extraction as well.

\subsection{Convolutional Neural Networks(CNNs)}
Convolutional neural networks are a type of neurobiologically inspired feed-forward artificial neural network which consist of multiple layers of neurons, with neurons in each layer collected into sets. At the input layer (where data is presented), these neuron sets map to small regions of input image. Deeper layers of the network can be composed of local or global pooling (fully-connected) layers which combine outputs of the neuron sets from previous layer. The pooling is typically achieved through convolution-like operations and hence the name . Figures \ref{fig:figure1} and \ref{fig:figure2} show an illustration of two popular CNNs. An attractive feature of CNNs is that the outputs of inner layers serve as a useful feature representation of the input -- a fact we exploit to obtain sketch features. The flexibility in the choice of layers, the local nature of operations in some parts of the network and crucially, the need to perform very little preprocessing on input images have all contributed to their impressive performance in advancing the state of the art for problems in computer vision and image processing\cite{krizhevsky2012imagenet}\cite{girshick2014rcnn}\cite{DonahueJVHZTD13}\cite{lecun1998gradient}. We have experimented with two well-known CNNs -- LeNet and Imagenet -- which we briefly describe below.

\textbf{Imagenet CNN}\cite{krizhevsky2012imagenet} (Figure \ref{fig:figure1}): This is a $8$-layer network (excluding input) with $5$ convolutional (locally connected) layers and $3$ fully connected layers. The output of the last (fully-connected) layer is connected to a $1000$-way softmax, thus producing a distribution of the $1000$ class labels. The number of class labels is a consequence of the number of image classes present in the Imagenet dataset\cite{imagenet} for which this CNN was designed. For our experiments, we utilize the pre-trained Imagenet trained on ILSVRC 2012 dataset containing $1.2$ million images, as described by Krizhevsky et al\cite{krizhevsky2012imagenet}. Our choice was motivated by the fact that features extracted from the Imagenet CNN have resulted in impressive performance across a variety of challenging computer vision problems\cite{girshick2014rcnn}\cite{deepimagenetfc7}. The sketch features are obtained by tapping the output of layer denoted as $fc_7$ of the pre-trained Imagenet CNN with the sketch as the input\cite{jia2014caffe} (see Figure \ref{fig:figure1}). As a result, we obtain a $4096$-dimensional feature vector corresponding to an input sketch.

\textbf{Lenet CNN}\cite{lecun1998gradient} (Figure \ref{fig:figure2}): This is a $7$-layer network (excluding input) with $2$ convolutional layers, $2$ subsampling layers, $2$ fully connected layers and a Gaussian connected layer with $10$ output classes. Our choice of LeNet was motivated by the fact that it is a CNN specially designed for recognizing handwritten digits, a problem whose modality (hand-generated) and nature of data (sparse binary shapes) resembles our sketch classification problem. We train a modified version of Lenet CNN\cite{jia2014caffe} using the MNIST handwritten digit dataset for $10000$ iterations at which point, accuracies are around $98\%$\cite{jia2014caffe}. The modified version of LeNet CNN can be seen in Figure \ref{fig:figure2} (only the first five layers are shown).  The sketch features are obtained by tapping the output of the inner-product layer of the trained Lenet CNN with the sketch as the input as in the case of Imagenet CNN. As a result, we obtain a $500$-dimensional feature vector corresponding to an input sketch.

\begin{figure*}[ht]
  \includegraphics[height=3.5cm,width=\textwidth]{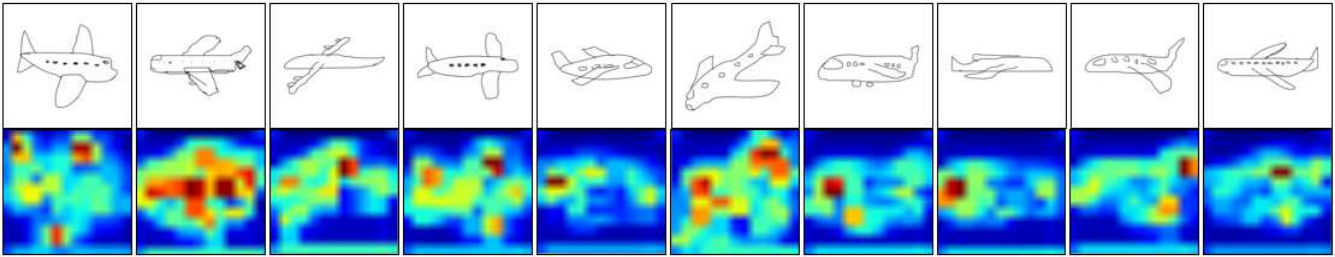}
  \caption{Sketches (top row) of category \texttt{airplane} and their corresponding $conv_5$ heat-maps (bottom row). The hotter (redder) the region, the more its relative significance in the feature representation. Regions of the heat-map corresponding to tail, nose and wingtips generally seem to be “hotter” than other parts of the object (airplane), implying their significance for representation and recognition.}
	\label{fig:heat-maps}
\end{figure*}

\subsection{Sketch data augmentation}
\label{sec:dataaug}
The benchmark database of Ros\'{a}lia et al.\cite{rosalia} contains only $56$ sketches per object category. To increase the number of sketches per category for classification, we perform data augmentation by applying geometric and morphological transformations to each sketch. Specifically, each sketch is initially subjected to image dilation (``thickening") using a $5 \times 5$ square structuring element. A number of transforms are applied to this thickened sketch --  mirroring (across vertical axis), rotation ($\pm 5,\pm 15$ degrees), systematic combinations of horizontal and vertical shifts ($\pm 5,\pm 15$ pixels), central zoom ($\pm 3\%,\pm 7 \%$ of image height). As a result, $30$ new sketches are generated per original sketch. The data augmentation procedure results in $30 \times 56 = 1680$ sketches per category, for a total of $1680 \times 160 = 268,800$ sketches across $160$ categories.

In the context of the sketches being processed by CNNs, we would also like to point out the reason for sketch dilation. As each sketch gets processed by deeper layers of the CNN, fine details tend to get eliminated. To minimize the impact of detail loss, the sketches from the database are subjected to thickening (dilation) which we believe helps preserve detail better.

The extracted training features are passed to a multi-class linear Support Vector Machine(SVM) classifier\cite{liblinear}. In the next section, we describe details related to training and evaluation of the sketch recognition system.

\section{Evaluation}
\label{sec:eval}

\begin{figure}
\centering
	\includegraphics[width=0.48\textwidth]{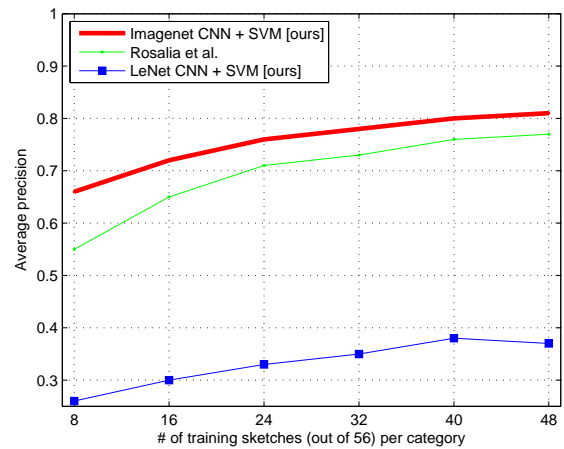}
	\caption{Classification Results}
	\label{fig:primality-trend-length}	
 \end{figure}

For the purpose of comparable evaluation, we utilize the same methodology and test set as that of Ros\'{a}lia et al.\cite{rosalia}. To begin with, our curated dataset contains $160$ sketch categories, each containing $56$ sketches. Since some of the categories contain more than $56$ sketches, we randomly select a $56$-sized subset from these categories. The number of sketches utilized for training is progressively increased, starting from $8$ sketches per category in steps of $8$ up to $48$ of the $56$ sketches. The rest of the sketches are utilized for testing. Note that for training, data augmented variants (Section \ref{sec:dataaug}) of each sketch are used while testing is done only on the original sketch subjected to dilation and not on the data augmented variants. As an example, when $32$ of the original sketches are used, the actual number of training sketches is $32 \times 30 \times 160 = 153,600$ while the number of test sketches is $24 \times 160 = 3840$. The entire data is randomly shuffled thrice and for each shuffle, it is split according to one of the training and testing splits mentioned above. For each shuffle, precision is calculated over test data and the $3$ precision values obtained for each shuffle are averaged. This procedure of shuffling thrice and computing average precision is repeated for each of the $8$ train/test splits considered. The results can be seen in Figure \ref{fig:primality-trend-length}. Our Imagenet CNN feature-based recognition system consistently outperforms the best results of Ros\'{a}lia et al. -- the improvement in precision ranges from $3\%$ to $11\%$. It must be conceded that the performance of Ros\'{a}lia et al. is based on unaugmented data. However, they employ Fisher vector feature representation whose dimensions\footnote{The Fisher vector feature dimension has been estimated based on the description of feature extraction process by Ros\'{a}lia et al.\cite{rosalia}.} are larger than our $4096$-dimensional feature vectors by a factor of about $10$. This would result in prohibitively large memory requirements and training times. In contrast, our feature vectors are faster to process. Therefore, they are also an ideal candidate for related applications such as sketch-based image retrieval\cite{Hu:2013:PEG:2479988.2480107}.

Figure \ref{fig:primality-trend-length} also shows that the results using Lenet CNN features are quite inferior. In hindsight, we realized that the sketches of a category exhibit more variety compared to handwritten digits. The LeNet CNN is relatively simpler compared to Imagenet CNN and consequently, not as powerful in capturing the entire gamut of $160$ sketch categories. While strategies such as modifying the network and re-training using sketch data are possible, it might be more fruitful to explore networks similar in spirit to Imagenet CNN, which demonstrate better performance. 

Although the focus of the paper has been on the recognition framework, fine-grained analysis of results, similar to that presented by Eitz\cite{eitz} et al. and  Ros\'{a}lia et al.\cite{rosalia}, can reveal interesting insights. As a preliminary step in this direction, we explore how deep features can help identify importance of crucial attributes (e.g. object-parts) of the sketched objects. Since attributes such as parts are spatial in nature, such features need to be extracted from CNN layers which preserve spatial information. $conv_5$ in Imagenet CNN is one such layer\cite{krizhevsky2012imagenet}. Therefore, we first extract $conv_5$ features-map (also referred to as ``heat-map") of the Imagenet CNN for each input sketch image. The feature-map can be thought of as an image whose intensity at a location represents the degree to which filters in $conv_5$ layer fire for the given input image\cite{agrawal14analyzing}. Such heat-maps can be used for a part-based analysis of the sketches. For example, ``heat-maps" in Figure \ref{fig:heat-maps} seem to suggest the importance of tail , nose and wingtips for sketches belonging to the category \texttt{airplane}. Therefore, relative to other parts such as windows, fuselage etc., these parts probably contribute crucially to the general representation and consequently, recognition of sketches belonging to \texttt{airplane} category.  Heat-maps for other categories can be viewed at \url{http://val.serc.iisc.ernet.in/sketchrec/sr.html} \hspace{1.5mm}.


\section{Conclusion and Future Work}
\label{sec:conclusion}

In this paper, we have presented a deep-learning based framework for recognizing freehand sketches of object categories. Our choice of features ensures the average precision improves by $3\% - 11\%$ over the existing state-of-the-art results. Our main novelty -- ``deep sketch features" -- has the potential to improve performance for the related problem of sketch-based image retrieval\cite{Hu:2013:PEG:2479988.2480107}. In addition, we provide a preliminary glimpse of how such features can help identify relative importance of crucial attributes (e.g. object-parts) in the sketched objects. More generally, we hope that our work will spur interest in analysis of sketches utilizing cutting-edge tools of the deep learning methodology.

A number of directions exist for future work. One obvious direction, currently in progress, is to compare the performance across a larger set of Convolutional Neural Networks and analyze the effect of tapping features from different layers of such networks. Another direction would be to explore the use of temporal stroke information available for each sketch in the database. It would be interesting to explore classifiers similar in spirit to Hidden Markov Models(HMM)s which can exploit such information for better sketch recognition. 

\bibliographystyle{IEEEbib}
\bibliography{refs}

\vfill

\end{document}